\documentclass[10pt,twocolumn,letterpaper]{article}

\usepackage[pagenumbers]{cvpr} 
\usepackage{booktabs}

\definecolor{cvprblue}{rgb}{0.21,0.49,0.74}
\usepackage[pagebackref,breaklinks,colorlinks,allcolors=cvprblue]{hyperref}

\def\paperID{147} 
\def\confName{CVPR}
\def\confYear{2026}

\newcommand{\dataset}{DesignPref}

\title{\dataset: Capturing Personal Preferences in Visual Design Generation}

\author{Yi-Hao Peng\\
Carnegie Mellon University\\
{\tt\small yihaop@cs.cmu.edu}
\and
Jeffrey P. Bigham\\
Carnegie Mellon University\\
{\tt\small jbigham@cs.cmu.edu}
\and
Jason Wu\\
Apple\\
{\tt\small jason\_wu8@apple.com}
}

\usepackage{graphicx}
\usepackage{subcaption}

\usepackage{booktabs}
\usepackage{siunitx}
\usepackage{array}

\usepackage{multirow, tabularx}
\usepackage{tcolorbox}

\newcolumntype{L}[1]{>{\raggedright\arraybackslash}p{#1}}

\begin{document}

\maketitle

\begin{abstract}
Generative models, such as large language models and text-to-image diffusion models, are increasingly used to create visual designs like user interfaces (UIs) and presentation slides.
Finetuning and benchmarking these generative models have often relied on datasets of human-annotated design preferences. Yet, due to the subjective and highly personalized nature of visual design, preference varies widely among individuals. 
In this paper, we study this problem by introducing DesignPref, a dataset of 12k pairwise comparisons of UI design generation annotated by 20 professional designers with multi-level preference ratings. We found that among trained designers, substantial levels of disagreement exist (Krippendorff’s alpha = 0.25 for binary preferences). 
Natural language rationales provided by these designers indicate that disagreements stem from differing perceptions of various design aspect importance and individual preferences.
With DesignPref, we demonstrate that traditional majority-voting methods for training aggregated ``judge'' models often do not accurately reflect individual preferences. To address this challenge, we investigate multiple personalization strategies, particularly fine-tuning or incorporating designer-specific annotations into RAG pipelines. Our results show that personalized models consistently outperform aggregated baseline models in predicting individual designers' preferences, even when using 20 times fewer examples.
Our work provides the first dataset to study personalized visual design evaluation and support future research into modeling individual design taste.
\end{abstract}

\section{Introduction}
\label{sec:intro}

\begin{quote}
\small
\emph{``Beauty is no quality in things themselves: It exists merely in the mind which contemplates them; and each mind perceives a different beauty.''}--- \textsc{David Hume}
\end{quote}
\noindent
Modern generative models can now support visual design tasks across UI layout and presentation slides composition. Design quality matters at scale: a classic survey reports that about half of an application’s code and a similar share of development time focus on the user interface~\citep{myers1992survey}. With advances in large language, vision-language and multimodal models (LLMs/VLMs/LMMs), recent work begins to assess visual design quality automatically. Across designs, methods pair compiler or runtime diagnostics with text–image alignment to estimate intent match~\citep{wu2024uicoder}, train scorers that judge design screenshots against written goals and align with professional judgments~\citep{wu2024uiclip, yun2025designlab}, and add region‑level critiques for design to turn scores into design suggestion~\citep{duan2024uicrit}. These works suggest that model‑based evaluation can potentially guide design workflows at scale.

Despite rapid progress, off‑the‑shelf evaluators do not yet deliver accurate personalized assessment of visual design.
Most preference data is collected for the purpose of aggregation in benchmark~\citep{zheng2023judging,webdev_arena_2025} or training settings such as RLHF and DPO~\citep{ouyang2022training,rafailov2023direct}.
Yet, previous work has shown that visual preference can vary widely among individuals and be poorly approximated by globally aggregated data~\citep{kirstain2023pick,lee2019image,goree2023correct}. 
This is especially true for UI design, which has been shown to vary based on aesthetic taste, user needs~\cite{wobbrock2011ability,peng2019personaltouch}, cultural background~\cite{reinecke2014quantifying}, and design trends~\cite{goree2021investigating,wu2023webui,li2025waybackui}.
Prior work has attempted to apply rubric-guided ratings and rankings to collect designer feedback~\cite{duan2024uicrit,luther2014crowdcrit} but found that subjectivity within the guidelines and differing preferences among designers led to a noisy learning signal for model training~\cite{duan2024uicrit,wu2024uiclip}.

To better support visual design personalization, we introduce \textsc{\dataset}, a dataset of 12,000 pairwise comparisons over generative UI designs, each with binary and four-class preference ratings from one of 20 professional graphic designers. Analysis reveals substantial disagreement across trained designers (Krippendorff's alpha = 0.25 for binary preferences). 28.5\% of comparisons have 96\% or more pair-wise disagreement rate.
We further collect the rationale on top of pairs that raise highly divergent preferences.
Our results show that designers can have different preferences on color contrast (dark or bright), how dense the screen should be (detailed vs simple), or how strongly hierarchy is marked.
To show the utility of individual design preference data in improving model performance, we conduct several experiments.
We show that finetuning an existing design assessment model~\cite{wu2024uiclip} on a designer's individual labels leads to better preference prediction accuracy and rating correlation than the same model trained on a 20 times larger aggregated dataset.
We conduct a similar experiment with an existing RAG-based UI assessment system~\cite{duan2024uicrit} and show similar improvements when retrieving from a smaller, personalized pool of examples.

In summary, our work makes three contributions towards personalizing design assessment. First, we introduce \textsc{DesignPref}, a benchmark of 12k identity‑linked, designer‑authored pairwise comparisons with rationales that supports research on personalized evaluation of visual UI design. Second, we conduct an analysis of this dataset where we quantify the prevalence of inter-rater disagreements and perform a coding-based analysis of designers' rationales to better understand the reasons behind divergence. Finally, we conduct a series of modeling experiments that demonstrate a smaller amount of personalized data leads to better preference prediction than a 20 times larger aggregated dataset in a finetuning and RAG setting.



\section{Related Work}
\label{sec:related_work}
To contextualize our work, we review previous research on automated UI design assessment, visual model-as-a-judge systems, and personalized image assessment.
 
\paragraph{Automated Assessment of UI Design.} 
Recent advances in LLMs/LMMs make automatic UI design assessment feasible. Early work prompts GPT‑4 with written heuristics; expert designers report helpful suggestions yet low stability across revisions~\cite{duan2024generating}. Followed with that, researchers released UICrit, which is a dataset contains expert critiques with region boxes and quality ratings and showed that few‑shot and visual prompts raise LLM feedback quality~\cite{duan2024uicrit}.
Recent benchmark treated LMM as unified UI judges and found only moderate alignment with human preferences across absolute and pairwise evaluations~\cite{wu2024uiclip,luera2025mllm}. 
To make design quality modeling effective, recent UI design autorater adopts UI- and quality-aware signals instead of generic vision-language objectives. \textsc{UIClip} introduced a design quality-oriented encoder that scores prompt relevance and design quality and matched professional designer preferences better than off-the-shelf LMM-as-a-judge while also producing design suggestions~\cite{wu2024uiclip}. All these prior efforts focus on optimizing aggregated, universal visual design judges. Our work instead focuses on modeling personal judgment, where we retain rater identity and a four‑level intensity scale to make models better predict each designer’s choices.

\paragraph{Model-as-a-Judge for Generative Visuals.}
Work on visual generation often relies on model judges that capture \emph{population} preferences and then steer and improve generators with those rewards. Two lines of progress set the stage. 
First, automatic evaluators replace generic proxies with task-aware checks: TIFA asks a VQA system questions about the prompt to measure faithfulness and shows stronger agreement with humans than CLIP-style metrics~\citep{hu2023tifa}. 
Second, preference-trained reward models turn large-scale human choices into scalar scores: ImageReward, Pick-a-Pic, and HPSv2/v3 are examples of models trained using expert pairwise comparisons and votes~\citep{xu2023imagereward,kirstain2023pick,ma2025hpsv3}. These types of automated judges can enable preference optimization for generative models~\citep{wallace2024diffusion,karthik2025scalable} without hand‑crafted rewards.
Most of these works aim to learn a single global utility and collapse diverse tastes into an average score. \textsc{DesignPref} complements this setting by keeping rater identity and a four-level confidence label on each comparison, so a judge can learn whose preference to predict and at what strength. This personalized view matters for design-centric visuals where trained professionals still disagree at a high rate, and where downstream tuning can potentially benefit from identity-aware targets.

\paragraph{Personalized Image Assessment.}
Personalized visual preference research models user‑specific taste instead of a population mean. The field first framed personalized aesthetics as a supervised problem with rater identity: Ren et al.\ introduced a task with AMT rater IDs and owner‑rated albums and showed that content and attribute cues predict user‑level deviations from a generic scorer~\citep{ren2017personalized}. PARA broadened scale and subject coverage and added rich user attributes, which enables identity‑conditioned prediction~\citep{yang2022para}.
Personalized image quality assessment (IQA) work has also trained models for specific age groups and individuals~\citep{wang2023agespecificiqa,cherepkova2024contrast} to improve traditional metrics such as NIMA, MUSIQ, and CLIP-IQA~\citep{talebi2018nima,ke2021musiq,wang2023clipiqa}
In our paper, we seek to extend this research to UI design assessment, where individual preferences could be influenced by a range of factors such as aesthetic taste, accessibility needs~\cite{wobbrock2011ability,peng2025morae,peng2021slidecho}, cultural background~\cite{reinecke2014quantifying}, and evolving design trends~\cite{goree2021investigating,li2025waybackui}.
To this end, we contribute a dataset of identify-linked designer preferences and reasoning which we show can personalize VLMs through fine-tuning and retrieval-augmented generation (RAG) approaches.


\begin{figure}[t]
    \centering
    \includegraphics[width=0.45\textwidth]{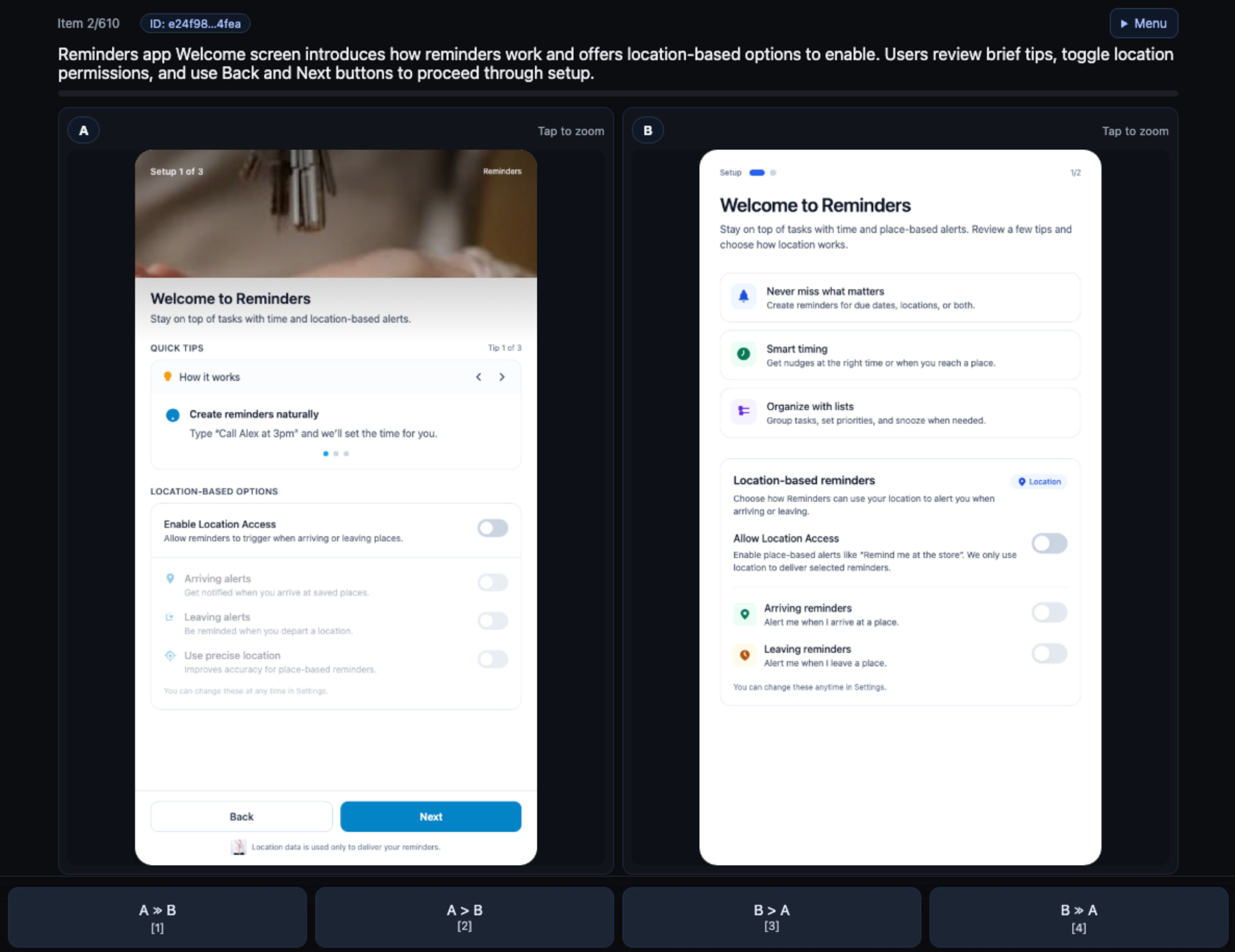}
    \caption{Our annotation interface includes a prompt that defines the design task. Two variants generated from the prompt appear side by side. The rater selects one of four preference options.}
    \label{fig:annot_ui}
\end{figure}

\section{DesignPref Dataset}
Our work aims to understand and model how individual design preferences arise for UIs.
We introduce DesignPref, a dataset that captures personalized preferences for generative visual design. We focus on the domain of mobile UI, where A/B testing and comparisons are standard practice~\citep{king2017designing,delamaro1996integration}.

\subsection{Dataset Construction}
\label{sec:designpref}
Existing UI datasets do not include annotations for preference personalization, yet they provide screenshots, text descriptions, and other metadata that we reuse to build controlled pairs and collect judgments from individual designers~\citep{wang2021screen2words,bunian2021vins,wu2024uiclip,duan2024uicrit}. The resulting dataset, methods, and findings can be potentially extended to other graphic design domains such as presentation slides~\cite{ge2025autopresent,peng2023slide,yun2025designlab,peng2022diffscriber} and posters~\cite{hsu2023posterlayout,hsu2025postero}.
We choose GPT-5 and Gemini-2.5-Pro, two strong-performing language models from WebDev Arena~\citep{webdev_arena_2025}, as base generators. Each model produces executable UI code (HTML, CSS, and JavaScript) to render mobile screens. Our study measures how people judge visual design quality and how those judgments vary across individuals, capturing shared criteria as well as personal taste.

\subsubsection{Design Generation}
We sampled 100 human-written screen descriptions from existing dataset~\citep{wang2021screen2words} and extended them to a more complete screen descriptions using established method~\cite{peng2024dreamstruct} as seed prompts. To ensure coverage across UI types, we stratified samples in different categories from prior defined taxonomy \citep{leiva2020enrico}. For each description, each model produced four variants with a high temperature to increase diversity. We used similar standard Design Arena~\cite{design_arena_2025} prompts to elicit high-quality, renderable UI code and included explicit quality checks in the instructions. In total, our dataset contains 400 generative designs. Each prompt yields 6 within-prompt pairs, which gives 600 unique comparison pairs across all generation.

\subsubsection{Data Annotation Protocol}
We recruited 20 designers via a university mailing list and online platforms. 
Each designer who participated in our study had at least one year of professional UI design experience.
We designed our data annotation task to take approximately 90 minutes to complete. Designers who participated in our study were compensated \$15.55 an hour, and our data collection protocol was approved by our institution's IRB.

Each designer viewed a pair of UI screens $(A,B)$ in each trial. The rater chose both direction and strength of preference on a four‑point Likert‑scale (Fig.~\ref{fig:annot_ui}): ``A much better than B'' (A$\gg$B), ``A better than B'' (A$>$B), ``B better than A'' (B$>$A) or ``B much better than A'' (B$\gg$A). 
We chose an even-numbered scale (i.e., no neutral option) to encourage raters to express even subtle preferences.
The protocol follows recent works that use fine‑grained preference options for model alignment and evaluation~\citep{touvron2023llama2,song2024veriscore,wang2024lrhp}.  
We linked every label to a persistent rater identifier to support per‑rater models and personalization analyses, consistent with work that estimates judge‑specific reliability in crowdsourcing settings~\citep{chen2013pairwise,caron2012gbt,dawid1979ml,raykar2010crowds}.

\subsection{Dataset Analysis}
\label{sec:dataset_analysis}
Each designer rated 600 UI pairs under two label granularities: a binary direction label that records which variant wins, and a four-way label that encodes preference strength. Group-level reliability remains modest in both schemes. For binary preferences, the mean pairwise agreement is $0.624$, and both mean Cohen's $\kappa$ and multi-rater Krippendorff's $\alpha$ equal $0.248$. For the four-way labels, the corresponding values drop to $0.386$ for agreement, $0.114$ for $\kappa$, and $0.104$ for $\alpha$. These values confirm substantial disagreement among trained designers and align with prior designer studies~\citep{wu2024uiclip,duan2024uicrit}. Direction and four-way $\kappa$ correlate at $r=0.60$, so designer pairs who agree on winners also tend to agree on strength, although strength judgments add additional noise.

\begin{figure}[t]
    \centering
    \includegraphics[width=0.45\textwidth]{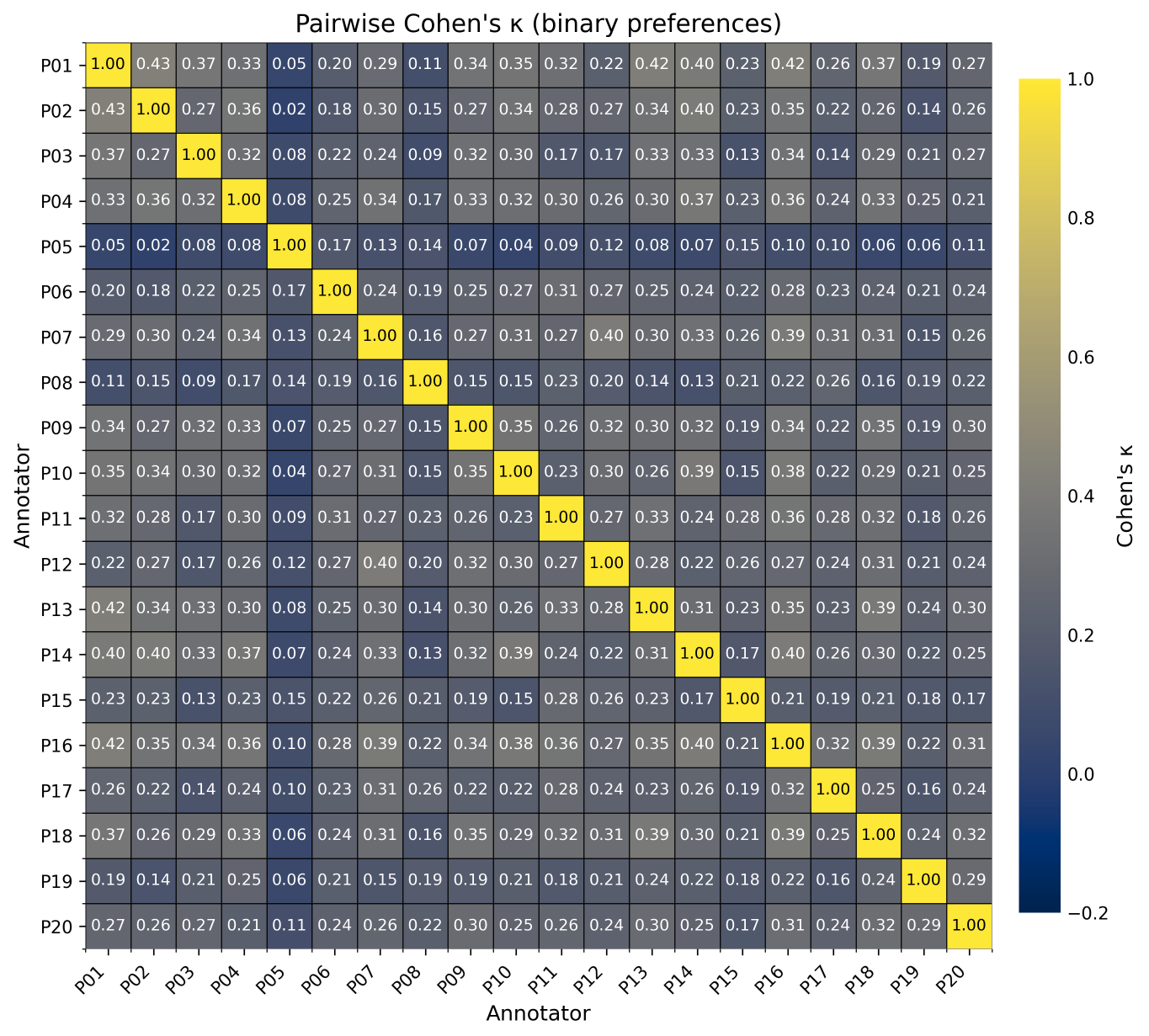}
    \caption{Cohen’s kappa agreement for pairwise binary preferences across designers.}
    \label{fig:kappa}
\end{figure}

Label-use patterns help explain the reliability gap between direction and four-way evaluation. Class frequencies for the four-way choices are A$>$B $36.97\%$, B$>$A $38.57\%$, A$\gg$B $11.82\%$, and B$\gg$A $12.64\%$. When aggregated by strength, $75.54\%$ of judgments use the middle options (A$>$B or B$>$A), and $24.46\%$ use the extreme options (A$\gg$B or B$\gg$A). Usage varies markedly across designers: the middle-option share has mean $75.38\%$ with values from $6.00\%$ to $99.50\%$, and the extreme-option share has mean $24.41\%$ with values from $0.17\%$ to $93.67\%$. The distribution suggests frequent reliance on slight preferences and highly heterogeneous thresholds for strong claims.

\paragraph{Effect of Screen Type.}
We hypothesized that the type of UI screen may have contributed to designers' preference distribution, as some screens (e.g., social media feeds) may surface more variations than others (e.g., login screens).
To analyze the agreement level on different UI screen types, we group items by screen types defined by prior taxonomy~\cite{leiva2020enrico} and measure direction consensus with the per-item mean pairwise agreement and the entropy of binary votes, then average within each type.
\emph{Form} screens show the strongest consensus, with mean pairwise agreement of 72.11\% and the highest share of extreme choices 32.71\%.
\emph{List}, \emph{Dialer}, and \emph{Search} screens follow with agreement around 66\%–68\%.
Designers disagree most on \emph{Terms} pages, \emph{Camera} interfaces, \emph{Media Player} screens, and \emph{Login} flows, all of which stay near 54\%–57\% agreement.
Across the 20 types, mean agreement and the share of extreme judgments correlate strongly ($r=0.774$), so screen types that yield clearer winners also elicit stronger preference consensus.
Overall, structured layouts with clear primary actions (e.g., forms, search pages) support more unified and decisive preferences, whereas visually complex or multifunctional screens lead to more diffuse judgments.

\subsection{Understanding Rationale Behind Preferences}
We explore how designers explain their preferences with a rationale annotation study. We invited the same 20 designers from the preference study; 13 participated. The interface resembled the original one but added a text box under each selected pair. For each chosen comparison, designers saw their earlier choice and wrote a brief rationale. To understand disagreements, we sampled highly contested pairs with majority margin $|\text{choose A}\% - \text{choose B}\%|\le 10\%$, which corresponds to splits of 55\% versus 45\% or closer. This process yielded 106 design pairs. Across these pairs, designers wrote 1{,}378 rationales, with about 3 hours of work per participant (we paid \$15.55 per hour).

\begin{figure}[t]
    \centering
    \includegraphics[width=0.46\textwidth]{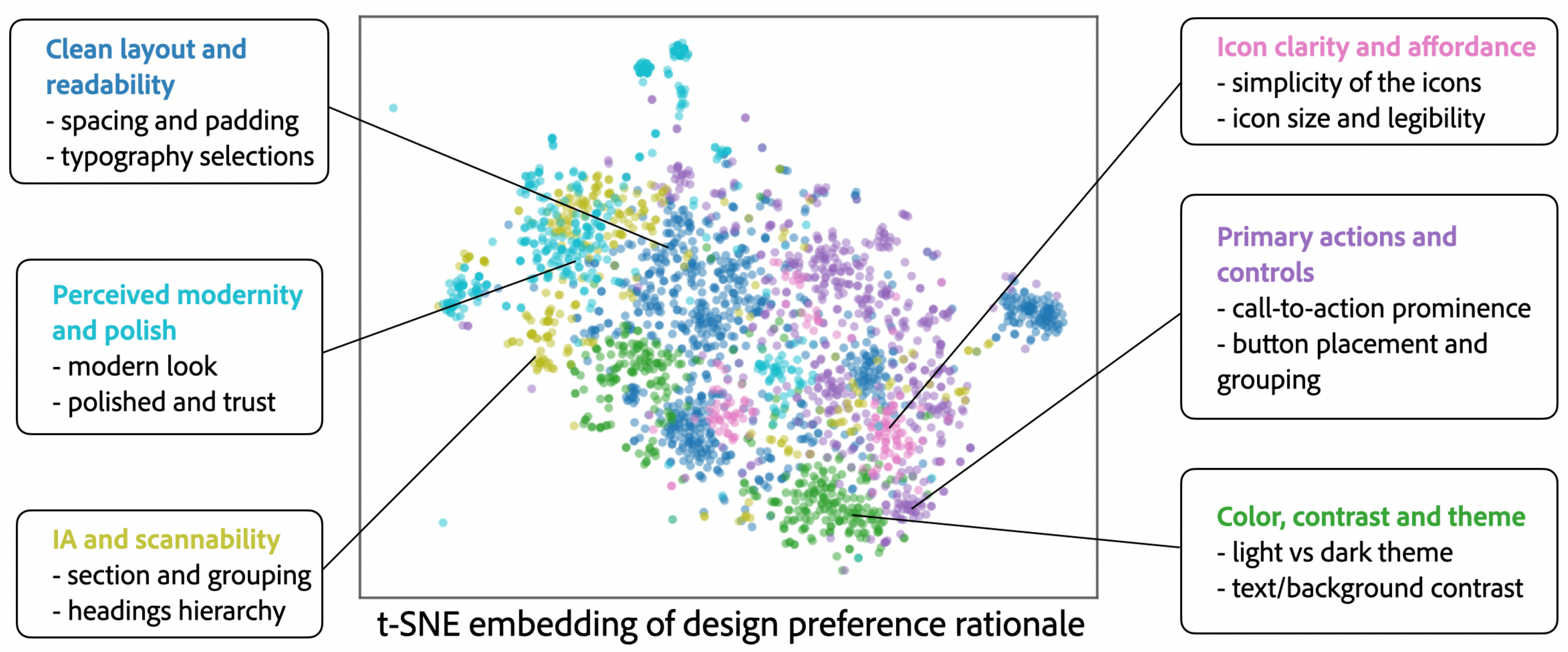}
    \caption{Embedding clusters for preference rationale themes.}
    \label{fig:rationale_tsne}
\end{figure}

\begin{figure*}[htbp]
    \centering
    \begin{subfigure}[htbp]{0.475\textwidth}
        \centering
        \includegraphics[width=\linewidth]{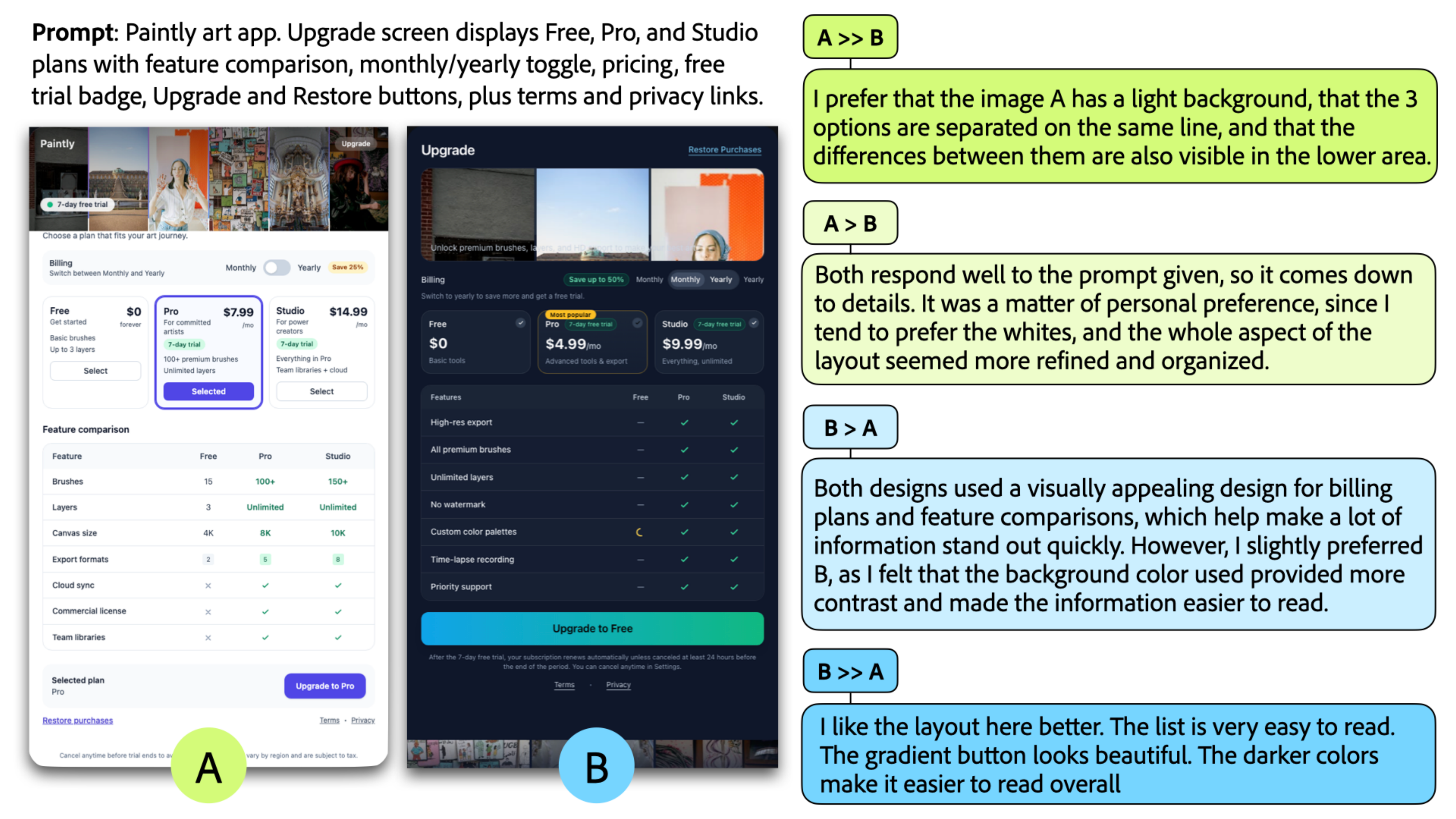}
        \caption{In the upgrade screen, designers diverge mainly on preference for light versus dark themes and contrast.}
        \label{fig:div_exmp_1}
    \end{subfigure}
    \hfill
    \begin{subfigure}[htbp]{0.475\textwidth}
        \centering
        \includegraphics[width=\linewidth]{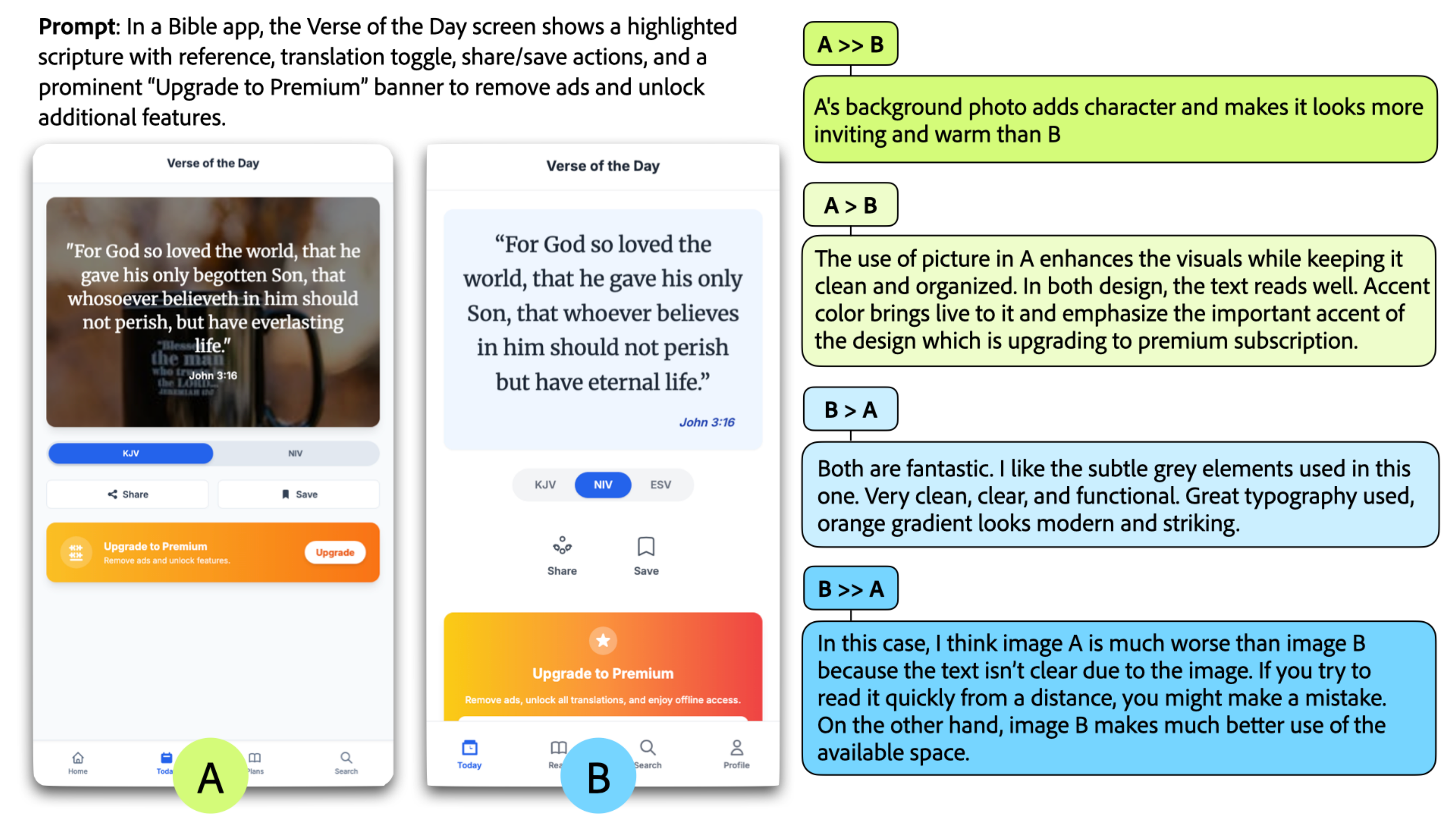}
        \caption{ In the Bible app screen, designers diverge between decorative imagery and a more utility-focused layout.}
        \label{fig:div_exmp_2}
    \end{subfigure}
    \caption{Rationales behind divergent preferences. Each pair shows an example where half of the designers chose A and half chose B.}
    \label{fig:div_exmp}
\end{figure*}

\subsubsection{Common Themes in Preference Rationale}
We characterize common patterns in the rationales with a two-stage, embedding-based clustering pipeline that follows recent work on automatic and human-centered thematic analysis~\cite{qiao2025thematic,wang2025lata,beeferman2023feedbackmap,peng2021say}. We split each rationale into sentences and treat each sentence as a basic reasoning unit. A transformer-based sentence encoder maps each unit into a high-dimensional vector space~\cite{reimers2019sentence}. We cluster these vectors into fine-grained groups and ask GPT-5 to assign each group a neutral label, short summary, and key concerns. A second GPT-5 pass merges related groups into broader themes.
Our analysis of 1{,}378 written rationales surfaces six recurrent themes (Figure~\ref{fig:rationale_tsne}), listed in order of how often designers mention them: \emph{clean layout and readability}, \emph{primary actions and controls}, \emph{perceived modernity and polish}, \emph{color, contrast, and theme}, \emph{information architecture (IA) and scannability}, and \emph{icon clarity and affordance}. 
\emph{Clean layout and readability} covers comments about generous spacing, clear alignment, and larger type.
\emph{Primary actions and controls} reflects a preference for obvious, reachable call-to-actions that stand apart from secondary options.
When screens already satisfy basic readability, \emph{perceived modernity and polish} often decides between functionally similar variants. Designers often favored crisper typography, smoother spacing, and cohesive accent colors and describe the preferred screen as more ``modern'' or ``professional''. 
\emph{Color, contrast, and theme} focuses on background tone and palette cohesion, where contrast helps key controls stand out. 
\emph{Information architecture and scannability} emphasizes clear sections and hierarchy, such as plan comparison screens with card-based layouts. 
\emph{Icon clarity and affordance} captures choices on dialer and camera screens, where fewer, larger icons with higher contrast and short labels make each control easier to interpret.

\subsubsection{Divergence between Preference Rationale}
While the previous analysis surfaces what designers generally agree on, contested pairs reveal structured disagreement in how designers weigh the same concerns. We focus on the selected contested pairs and treat each pair as a unit of analysis. For each pair, we give GPT-5 the full set of rationales from designers on each side. The model describes the main design trade-off in that pair and the main reasons respondents favor each version. A second GPT-5 pass groups related trade-offs into broader divergent themes.
The analysis highlights four recurrent themes. The first theme concerns \emph{information density}. Some designers prefer calm layouts with a short list of options or plans, while others favor denser pricing or support screens that surface more channels or help paths in one view. The second theme concerns \emph{visual style and tone}. Across login, consent, and fintech screens, some rationales emphasize light, neutral themes that feel legible and trustworthy. Others praise darker or more saturated palettes as more premium, expressive, or more institutional language that signals seriousness (Figure~\ref{fig:div_exmp_1}). The third theme concerns \emph{decorative imagery versus focused utility}. On some screens (e.g., media page), one side values large photos, gradients, or illustrations that add warmth and brand expression, while the other side prefers restrained, text-led layouts where icons and controls keep attention on the task or main goal of the app (Figure~\ref{fig:div_exmp_2}). The fourth theme concerns how strongly the interface foregrounds \emph{actions and task scope}. Task-first rationales prefer a single high-contrast primary button and minimal chrome, such as camera screens with one clear capture control and little status text. Feature-first rationales endorse richer layouts that expose more modes, filters, alerts, or secondary actions and describe very sparse screens as underpowered.

\section{Modeling Approaches}
\label{sec:modeling}
In this section, we describe approaches for incorporating DesignPref into machine learning models and pipelines to improve their performance in predicting personalized design preferences.
We focused on two example setups that have been used by previous work to assess UI design quality: i) finetuning a CLIP-style dual-encoder VLM~\cite{wu2024uiclip}, and ii) improving a decoder VLM/LMM with a multimodal RAG pipeline~\cite{duan2024uicrit}.

\paragraph{Data Processing.}
We took steps to prepare the DesignPref for our modeling experiments. First, we processed our collected data into a consistent format that could be used for fine-tuning and RAG pipelines.
\begin{equation}
\mathcal{D}_{\mathrm{pref}}
\;=\;
\{\, (t,\, I^{A},\, I^{B}, c) \;:\; c \in \{-2, -1, 1, 2\} \}
\label{eq:preference-data}
\end{equation}
In the above equation, we construct tuples for every labeled example consisting of the text $t$, image choices $I^A$ and $I^B$, and the recorded choice $c$ mapped to an integer value.

In addition, we created stratified data splits for our modeling experiments.
We chose to use \(60\%\) train, \(20\%\) validation, and \(20\%\) test split proportions to provide a large enough sample size for measuring performance metrics.
Our splits were constructed so that all pairs containing the a given screen ID belonged to the same split, preventing data leakage.
Across the entire dataset, this resulted in 7200 samples for the training set and 2400 samples for both the validation and test splits.
For each designer, this resulted in 360 samples for their training split and 120 samples for the validation and test splits.

\subsection{CLIP Finetuning}
We show the utility of our preference data in improving a CLIP-style dual-encoder VLM, which has previously been used to assess UI design quality and relevance~\cite{wu2024uiclip,wu2024uicoder}.
The CLIP architecture contains two transformer-based encoders and previous work~\cite{wu2024uiclip,wu2024uicoder} has used it to score UI by computing the cosine similarity between its embedded screenshot and a natural language description of the screen, e.g., ``ui screenshot. well-designed. settings page of an e-reader.''

Previous work~\cite{wu2024uiclip} found that the original CLIP model does not perform well on UI design assessment, and the authors finetuned a derivative model, called UIClip, on a large dataset of webpages screenshots with varying levels of synthetically-induced design flaws to better calibrate the predicted score~\cite{wu2024uiclip}.
In our experiments, we used UIClip's released model and inference code as a starting point.
Specifically, because we wanted to isolate the effect of our collected dataset, we chose to use UIClip's pretraining checkpoint\footnote{\url{https://huggingface.co/biglab/uiclip_jitteredwebsites-2-224-paraphrased}}, which was not fine-tuned on the authors' previously collected human preferences. 
Our training approach largely followed the pairwise contrastive learning approach used by UIClip to finetune on preference pairs~\cite{wu2024uiclip}.
However, we incorporate additional granularity from our data and better support model personalization.

\subsubsection{Strength-aware Margin}
First, we altered the originally proposed pairwise contrastive loss to a formulation with a cost-sensitive margin~\cite{iranmehr2019cost}, which allowed us to incorporate the strength of designers' preferences.
We used the following loss function for finetuning UIClip on preference examples from our dataset.
\begin{equation}
\begin{aligned}
\mathcal{L}(s^A, s^B; \hat{m})
&= \max\left\{0,\ \hat{m} - (s^A - s^B) \right\}, \\
\hat{m} &=
\begin{cases}
m_1, & \text{if A $>$ B},\\
m_2, & \text{if A $\gg$ B}.
\end{cases}
\end{aligned}
\end{equation}
In the above equation, $\hat{m}$ is the margin used for loss computation, which depends on the ground truth label.
$s^A$ and $s^B$ represent the model's predicted scores for a preferred sample $A$ and rejected sample $B$, respectively. 

During model inference, the difference between predicted screens' scores is used to make a four-way classification prediction.

\begin{equation}
\hat y(s^A,s^B;\hat m)=
\begin{cases}
\text{A}>\text{B},   & 0 \le s^A - s^B < m_2,\\
\text{A}\gg\text{B}, & s^A - s^B \ge m_2.
\end{cases}
\end{equation}

Negating these values leads to the thresholds for the opposite preference (e.g., B is better than A).
\subsubsection{Model Personalization}
In addition to using a strength-aware margin, we trained separate models on their preference pairs to better reflect their individual preferences.
Due to the limited amount of training data from each participant (360 examples), we found that only unfreezing the last layer of UIClip during training reduced the risk of overfitting and led to better performance in our early experiments.
We also introduced other regularization measures, such as weight decay to further combat overfitting.
All designers' models were trained using the same set of hyperparameters and early stopping schedule.
This set of hyperparameters was initialized from previously published values~\cite{wu2024uiclip} and further refined through manual experimentation. Hyperparameter values can be found in the supplemental material.

\subsection{Retrieval-Augmented Generation}
Prior work reports that decoder-style LMMs perform poorly on zero-shot UI design assessment tasks~\cite{wu2024uiclip,duan2024generating,duan2024uicrit,duan2024visual}. We hypothesized that retrieval-augmented generation (RAG) with our dataset could improve performance. Rather than relying only on knowledge stored in model weights, our RAG setup retrieves a small set of labeled designer preference examples as few-shot context for each query. We build this retrieval pipeline on the design critique generation framework from UICrit~\cite{duan2024uicrit}, and we adapt the prompt to predict pairwise design preferences instead of natural language comments.

\begin{table*}[t]
\centering
\small
\setlength{\tabcolsep}{6pt}
\begin{tabular}{llccc}
\toprule
\textbf{Model Group} & \textbf{Setup} & \textbf{Preference Acc. (\%)} & \textbf{Four-way Acc. (\%)} & \textbf{SRCC} \\
\midrule
\multicolumn{5}{c}{\textbf{Personalized models}}\\
\midrule
\multirow{2}{*}{CLIP (UIClip)}
 & Personalized, strength-aware margin & \textbf{60.16} & 34.37 & \textbf{0.217} \\
 & Personalized, binary margin       & 57.24          & 28.89 & 0.180          \\
\midrule
\multirow{4}{*}{LMM judges, RAG}
 & GPT-5 (8-shot)                  & 58.89 & 38.53 & 0.211 \\
 & Gemini-2.5-Pro (8-shot)         & 56.53 & 22.35 & 0.170 \\
  & Qwen3-VL-235B-Thinking (8-shot)     & 56.83 & \textbf{42.67} & 0.132 \\
 & Qwen3-VL-30B-Thinking (8-shot)      & 56.20 & 40.41 & 0.128 \\

\midrule
\multicolumn{5}{c}{\textbf{Non-personalized models}}\\
\midrule
\multirow{2}{*}{CLIP (UIClip), pooled}
 & Pooled, strength-aware (personal training size = N) & 56.73 & 32.77 & 0.150 \\
 & Pooled, strength-aware (full training set = 20N)       & 57.45 & 41.23 & 0.196 \\
\midrule
\multirow{4}{*}{LMM judges, pooled RAG}
 & GPT-5 (8-shot)                  & 57.62 & 34.36 & 0.203 \\
 & Gemini-2.5-Pro (8-shot)         & 56.15 & 22.56 & 0.169 \\
 & Qwen3-VL-235B-Thinking (8-shot)     & 55.78 & 41.59 & 0.122 \\
 & Qwen3-VL-30B-Thinking (8-shot)      & 54.78 & 38.79 & 0.110 \\

\midrule
\multirow{6}{*}{Baselines}
 & UIClip (pretrained)             & 55.07 & 23.97 & 0.126 \\
 & OpenAI CLIP B/32                & 46.68 & 23.26 & -0.009 \\
 & GPT-5 (zero-shot)               & 57.70 & 31.35 & 0.216 \\
 & Gemini-2.5-Pro (zero-shot)      & 53.65 & 17.76 & 0.135 \\
  & Qwen3-VL-235B-Thinking (zero-shot)  & 55.36 & 22.09 & 0.131 \\
 & Qwen3-VL-30B-Thinking (zero-shot)   & 54.82 & 16.17 & 0.117 \\

\bottomrule
\end{tabular}
\caption{
Comparison of personalized and non-personalized models on each designer’s test split. We report binary accuracy, four-way accuracy, and Spearman's rank correlation coefficient (SRCC) between predictions and ground-truth preference labels. 
}
\label{tab:main}
\end{table*}

\subsubsection{Sample Indexing}
We largely followed UICrit's approach to generating index vectors using textual (screen description) and visual information (UI screenshot).
Following UICrit, we used an off-the-shelf text-embedding model, Sentence BERT~\cite{reimers2019sentence}, to generate a fixed-length embedding from the UI's natural language description.
We chose to use UIClip's visual encoder to generate embeddings for the screenshots, as opposed to the originally used CLIP model~\cite{duan2024uicrit}, since previous work suggested it had stronger UI screenshot retrieval performance~\cite{wu2024uiclip}.
We used these three outputs to generate two index vectors for each preference pair in the training set, by concatenating the text embedding with both potential orderings of the screenshot embeddings.
We used two index vectors for each training example because we wanted our retrieval metric to be invariant to the pair order.
The retrieval score for a training example was computed as the maximum of the cosine similarity between the query vector and two index vectors.

\subsubsection{Personalized and Pooled Inference}
During inference, a retrieval score was computed for each example in training set and the results from the top $k$ were injected into the input prompt. The prompt format can be found in the supplemental materials of this paper.
When running RAG with a single designer's labeled data, this process resulted in exactly $k$ retrieved samples, since each designer was only asked to label a pair once.
However, when running RAG on the entire dataset of $n$ designers' data, $k\cdot n$ samples are returned, since each designer labeled the exact same pairs, which result in the same retrieval scores.
In this case, we aggregate designers' labels by averaging all their assigned scores into a single number then rounding to the nearest integer.
If the final score rounded to 0, then we randomly chose between A $<$ B and A $>$ B for the few-shot example.



\section{Model Evaluations}
\label{sec:results}
To demonstrate the utility of our dataset in improving ML-based UI design assessment, we conducted experiments that compared data and model configurations with access to personalized data.
\subsection{Evaluation Procedure}
\paragraph{Tested Models.}
We tested several configurations and baselines to measure the effect of our data on CLIP finetuning and RAG.
For our CLIP finetuning experiments, we compared the following conditions:
\begin{enumerate}\setlength{\itemsep}{2pt}
\item \textbf{Personalized, strength-aware margin.} We fine-tuned a separate UIClip model for each designer using our strength-aware margin loss.
\item \textbf{Personalized, normal margin.} We fine-tuned a separate UIClip model for each designer using a standard binary contrastive margin loss.
\item \textbf{Pooled, full training set.} We fine-tuned a single UIClip model using the combined training data from all designers.
\item \textbf{Pooled, matched training size.} We first combined the training data from all designers then randomly selected 1/20 of the samples, to match the training size of a single designer's data.
\end{enumerate}

\noindent We tested our RAG pipeline with four LMMs: GPT-5~\cite{openai_gpt5_system_card_2025}, Gemini-2.5-Pro~\cite{google_gemini25_pro_model_card_2025}, Qwen3-VL-235B-Thinking~\cite{alibaba_qwen3_technical_report_2025}, and Qwen3-VL-30B-Thinking.
At the time of our experiments, GPT-5 and Gemini-2.5-Pro were the top-performing proprietary VLMs, and Qwen3-VL-235B and Qwen3-VL-30B were the top-performing open source VLMs that could fit on a GPU server and consumer GPU.
For each RAG model, we varied the pool of data that the RAG system had access to: i) each designer's own training set (personalized) and ii) all designers' labels used our pooled inference strategy (non-personalized).
We set our RAG pipeline to retrieve 8 of the most relevant examples, following the best performing configuration found by previous work~\cite{duan2024uicrit}.

\paragraph{Performance Metrics.}
To measure model performance, we chose three metrics based on those that have previously been used to measure the alignment of automated scoring metrics with human ratings~\cite{talebi2018nima}: i) binary accuracy, ii) four-way accuracy, and iii) Spearman rank correlation coefficient (SRCC) on four-way ratings.
Metrics were computed individually on each designers' test split then averaged among 20 designers, i.e., macro-averaging.

\subsection{Results}
\label{sec:results}
Table~\ref{tab:main} shows the overall results. Our analysis focuses on two questions: how designer-specific supervision compares to pretrained or zero-shot judges, and whether personalization helps when compared to global models pool labels across designers.

\subsubsection{Personalized vs.\ pretrained and zero-shot judges}

Across backbones, designer-specific supervision improves alignment with individual preferences. The per-designer UIClip model with a strength-aware margin reaches 60.16\% binary accuracy and 0.217 SRCC, compared to 55.07\% and 0.126 for the frozen UIClip encoder and 46.68\% and $-0.009$ for CLIP B/32. A personalized model that uses a strength-aware loss performs better than the binary variant on all three metrics, suggesting that four-level labels provide useful signal when the loss exposes preference strength.

Personalized RAG with LMM on each designer's preference also shows various levels of improvements over zero-shot LMM judges. For GPT-5, personalization raises binary accuracy from 57.70\% to 58.89\% and four-way accuracy from 31.35\% to 38.53\% with similar SRCC. Personalization also helps both Qwen models gain about 1--2 points in binary accuracy and more than 20 points in four-way accuracy when retrieval conditions on each designer's own labels, while SRCC stays roughly unchanged. 
The two selected metrics show different aspects of the performance: four-way accuracy measures exact matches to discrete preference strengths, reflecting class imbalance and designers' differing thresholds between slight and strong choices. SRCC instead assesses how consistently model scores preserve overall rankings across the four levels. Given uneven distributions with most labels in the middle, and varied use of extreme ratings among designers, strength predictions remain noisier than binary winner predictions.

\subsubsection{Personalized vs.\ pooled models}

Per-designer models consistently bring better personalized prediction than global pooling. In UIClip, the per-designer strength-aware model achieves higher binary accuracy and SRCC than pooled four-class models, even when the pooled judge sees roughly twenty times more labels per designer. With a matched training set size, the personalized models improve all three metrics.
RAG experiments with LMM show similar pattern but with more marginal improvements. For GPT-5, Qwen-235B-Thinking, and Qwen-30B-Thinking, retrieval over each designer's own preference labels yields higher binary and four-way accuracy and equal or better SRCC than pooled RAG that averages labels across designers. Gemini-2.5-Pro behaves close to neutral under personalization, which suggests that the base model already encodes a strong consensus prior. Overall, identity-aware personalization helps both finetuned CLIP and few-shot LLM judges align with individual design preferences more closely than global pooling.


\section{Limitations}

Although our work suggests that personalization is important for design, we note several limitations and provide avenues for future improvement.
First, our dataset is limited to feedback from 20 designers.
In contrast to existing crowd-sourced design benchmarks~\cite{webdev_arena_2025,design_arena_2025}, we intentionally chose to focus on skilled (and compensated) designers, who we felt were more likely to understand and accurately apply design guidelines. Yet, a larger sample size might facilitate new types of analyses, such as the emergence of ``clusters'' of designers who might have similar design tastes, e.g., minimalism or Bauhaus school. Another limitation of our data is that it focused primarily on labeling noise resulting from inter-rater disagreement; although personal uncertainty might be another strong contributing factor. Anecdotally, we observed that designers in our study sometimes changed their own answer during decision process, which further suggests the difficulty of the task for ML models.

Our modeling experiments explored only a subset of possibilities enabled by DesignPref, and we specifically focused on replicating and improving two existing published examples~\cite{wu2024uiclip,duan2024uicrit}.
Applying personalization approaches led to the best-performing model (60.16\% binary accuracy), which is offers a similar level of performance as reward models for challenging tasks such as math and instruction following~\cite{lambert2025rewardbench,malik2025rewardbench2}.
Other modeling experiments, e.g., using preference data to finetune a decoder VLM, or using designers' critiques as reasoning traces are promising next steps for further improving performance, and we expect our dataset to enable additional future exploration.

Finally, our results show that, in our tested configurations, a smaller amount of personalized data could lead to better preference prediction than a 20 times larger aggregated dataset.
However, we did not attempt to study the amount of individually-collected data needed for effective model personalization.
Each designer in our study spent approximately 90 minutes to construct their dataset of 600 labels, a lengthy process for most practical applications.
A promising future direction is to study the scaling trend of personal preference data or apply related techniques to improve sample efficiency~\cite{talebi2018nima,ke2021musiq,wang2023clipiqa,yun2024taskvector}.


\section{Conclusion}

DesignPref establishes a benchmark for personalized visual design evaluation with identity-linked judgments. The dataset logs 12,000 pairwise UI comparisons from 20 professional designers with four-class strength labels, which enables per-designer modeling. Analyses show low cross-designer consensus and reveal the agreement varies among different designers and evaluated UI types. Per-designer UIClip models beat the best pooled judges on held-out pairs, showing the sample efficiency of personalized data. 
DesignPref shifts evaluation and alignment from a single global objective toward models that reflect individual taste and lays foundations for personalized design generation.


{
    \small
    \bibliographystyle{ieeenat_fullname}
    \bibliography{main}

@misc{webdev_arena_2025,
  author       = {{LMSYS Org.}},
  title        = {WebDev Arena Leaderboard},
  year         = {2025},
  howpublished = {\url{https://web.lmarena.ai/leaderboard}},
}

@misc{design_arena_2025,
  author       = {{Design Arena}},
  title        = {Design Arena},
  howpublished = {\url{https://www.designarena.ai/}},
  year         = {2025},
  note         = {Accessed: 2025-11-05}
}

@techreport{openai_gpt5_system_card_2025,
  title       = {GPT-5 System Card},
  author      = {{OpenAI}},
  institution = {OpenAI},
  year        = {2025},
  month       = aug,
  url         = {https://cdn.openai.com/gpt-5-system-card.pdf}
}

@article{alibaba_qwen3_technical_report_2025,
  title         = {Qwen3 Technical Report},
  author        = {Yang, An and Li, Anfeng and Yang, Baosong and others},
  journal       = {arXiv preprint arXiv:2505.09388},
  year          = {2025},
  eprint        = {2505.09388},
  archivePrefix = {arXiv},
  primaryClass  = {cs.CL},
  doi           = {10.48550/arXiv.2505.09388},
  url           = {https://arxiv.org/abs/2505.09388}
}

@inproceedings{goree2021investigating,
  title={Investigating the homogenization of web design: A mixed-methods approach},
  author={Goree, Samuel and Doosti, Bardia and Crandall, David and Su, Norman Makoto},
  booktitle={Proceedings of the 2021 CHI Conference on Human Factors in Computing Systems},
  pages={1--14},
  year={2021}
}

@inproceedings{peng2019personaltouch,
  title={Personaltouch: Improving touchscreen usability by personalizing accessibility settings based on individual user's touchscreen interaction},
  author={Peng, Yi-Hao and Lin, Muh-Tarng and Chen, Yi and Chen, TzuChuan and Ku, Pin Sung and Taele, Paul and Lim, Chin Guan and Chen, Mike Y},
  booktitle={Proceedings of the 2019 CHI Conference on Human Factors in Computing Systems},
  pages={1--11},
  year={2019}
}

@article{wobbrock2011ability,
  title={Ability-based design: Concept, principles and examples},
  author={Wobbrock, Jacob O and Kane, Shaun K and Gajos, Krzysztof Z and Harada, Susumu and Froehlich, Jon},
  journal={ACM Transactions on Accessible Computing (TACCESS)},
  volume={3},
  number={3},
  pages={1--27},
  year={2011},
  publisher={ACM New York, NY, USA}
}

@inproceedings{luther2014crowdcrit,
  author    = {Kurt Luther and
               Amy Pavel and
               Wei Wu and
               Jari{-}Lee Tolentino and
               Maneesh Agrawala and
               Bj{\"o}rn Hartmann and
               Steven P. Dow},
  title     = {CrowdCrit: crowdsourcing and aggregating visual design critique},
  booktitle = {Computer Supported Cooperative Work, {CSCW} '14, Baltimore, MD, USA,
               February 15--19, 2014, Companion Volume},
  pages     = {21--24},
  publisher = {ACM},
  year      = {2014},
  doi       = {10.1145/2556420.2556788},
  url       = {https://doi.org/10.1145/2556420.2556788}
}

@inproceedings{reinecke2014quantifying,
  title={Quantifying visual preferences around the world},
  author={Reinecke, Katharina and Gajos, Krzysztof Z},
  booktitle={Proceedings of the SIGCHI conference on human factors in computing systems},
  pages={11--20},
  year={2014}
}

@techreport{google_gemini25_pro_model_card_2025,
  title       = {Gemini 2.5 Pro Model Card},
  author      = {{Google DeepMind}},
  institution = {Google DeepMind},
  year        = {2025},
  month       = jun,
  note        = {Model card updated June 27, 2025},
  url         = {https://storage.googleapis.com/deepmind-media/Model-Cards/Gemini-2-5-Pro-Model-Card.pdf}
}

@article{reimers2019sentence,
  title={Sentence-bert: Sentence embeddings using siamese bert-networks},
  author={Reimers, Nils and Gurevych, Iryna},
  journal={arXiv preprint arXiv:1908.10084},
  year={2019}
}

@article{iranmehr2019cost,
  title={Cost-sensitive support vector machines},
  author={Iranmehr, Arya and Masnadi-Shirazi, Hamed and Vasconcelos, Nuno},
  journal={Neurocomputing},
  volume={343},
  pages={50--64},
  year={2019},
  publisher={Elsevier}
}

@inproceedings{wu2024uiclip,
  title={UIClip: a data-driven model for assessing user interface design},
  author={Wu, Jason and Peng, Yi-Hao and Li, Xin Yue Amanda and Swearngin, Amanda and Bigham, Jeffrey P and Nichols, Jeffrey},
  booktitle={Proceedings of the 37th Annual ACM Symposium on User Interface Software and Technology},
  pages={1--16},
  year={2024}
}

@article{wu2024uicoder,
  title={Uicoder: Finetuning large language models to generate user interface code through automated feedback},
  author={Wu, Jason and Schoop, Eldon and Leung, Alan and Barik, Titus and Bigham, Jeffrey P and Nichols, Jeffrey},
  journal={arXiv preprint arXiv:2406.07739},
  year={2024}
}

@inproceedings{duan2024uicrit,
  title={Uicrit: Enhancing automated design evaluation with a ui critique dataset},
  author={Duan, Peitong and Cheng, Chin-Yi and Li, Gang and Hartmann, Bjoern and Li, Yang},
  booktitle={Proceedings of the 37th Annual ACM Symposium on User Interface Software and Technology},
  pages={1--17},
  year={2024}
}

@inproceedings{myers1992survey,
  title={Survey on user interface programming},
  author={Myers, Brad A and Rosson, Mary Beth},
  booktitle={Proceedings of the SIGCHI conference on Human factors in computing systems},
  pages={195--202},
  year={1992}
}

@book{king2017designing,
  title={Designing with data: Improving the user experience with A/B testing},
  author={King, Rochelle and Churchill, Elizabeth F and Tan, Caitlin},
  year={2017},
  publisher={" O'Reilly Media, Inc."}
}

@inproceedings{delamaro1996integration,
  title={Integration testing using interface mutation},
  author={Delamaro, Marcio Eduardo and Maldonado, Jose Carlos and Mathur, Aditya P},
  booktitle={Proceedings of ISSRE'96: 7th International Symposium on Software Reliability Engineering},
  pages={112--121},
  year={1996},
  organization={IEEE}
}

@inproceedings{wang2021screen2words,
  title={Screen2words: Automatic mobile ui summarization with multimodal learning},
  author={Wang, Bryan and Li, Gang and Zhou, Xin and Chen, Zhourong and Grossman, Tovi and Li, Yang},
  booktitle={The 34th Annual ACM Symposium on User Interface Software and Technology},
  pages={498--510},
  year={2021}
}

@inproceedings{leiva2020enrico,
  title={Enrico: A dataset for topic modeling of mobile UI designs},
  author={Leiva, Luis A and Hota, Asutosh and Oulasvirta, Antti},
  booktitle={22nd International Conference on Human-Computer Interaction with Mobile Devices and Services},
  pages={1--4},
  year={2020}
}

@article{kirstain2023pick,
  title={Pick-a-pic: An open dataset of user preferences for text-to-image generation},
  author={Kirstain, Yuval and Polyak, Adam and Singer, Uriel and Matiana, Shahbuland and Penna, Joe and Levy, Omer},
  journal={Advances in neural information processing systems},
  volume={36},
  pages={36652--36663},
  year={2023}
}

@article{zheng2023judging,
  title={Judging llm-as-a-judge with mt-bench and chatbot arena},
  author={Zheng, Lianmin and Chiang, Wei-Lin and Sheng, Ying and Zhuang, Siyuan and Wu, Zhanghao and Zhuang, Yonghao and Lin, Zi and Li, Zhuohan and Li, Dacheng and Xing, Eric and others},
  journal={Advances in neural information processing systems},
  volume={36},
  pages={46595--46623},
  year={2023}
}

@article{ouyang2022training,
  title={Training language models to follow instructions with human feedback},
  author={Ouyang, Long and Wu, Jeffrey and Jiang, Xu and Almeida, Diogo and Wainwright, Carroll and Mishkin, Pamela and Zhang, Chong and Agarwal, Sandhini and Slama, Katarina and Ray, Alex and others},
  journal={Advances in neural information processing systems},
  volume={35},
  pages={27730--27744},
  year={2022}
}

@article{rafailov2023direct,
  title={Direct preference optimization: Your language model is secretly a reward model},
  author={Rafailov, Rafael and Sharma, Archit and Mitchell, Eric and Manning, Christopher D and Ermon, Stefano and Finn, Chelsea},
  journal={Advances in neural information processing systems},
  volume={36},
  pages={53728--53741},
  year={2023}
}

@article{xu2023imagereward,
  title={Imagereward: Learning and evaluating human preferences for text-to-image generation},
  author={Xu, Jiazheng and Liu, Xiao and Wu, Yuchen and Tong, Yuxuan and Li, Qinkai and Ding, Ming and Tang, Jie and Dong, Yuxiao},
  journal={Advances in Neural Information Processing Systems},
  volume={36},
  pages={15903--15935},
  year={2023}
}

@inproceedings{hu2023tifa,
  title={Tifa: Accurate and interpretable text-to-image faithfulness evaluation with question answering},
  author={Hu, Yushi and Liu, Benlin and Kasai, Jungo and Wang, Yizhong and Ostendorf, Mari and Krishna, Ranjay and Smith, Noah A},
  booktitle={Proceedings of the IEEE/CVF International Conference on Computer Vision},
  pages={20406--20417},
  year={2023}
}

@inproceedings{chen2013pairwise,
  title={Pairwise ranking aggregation in a crowdsourced setting},
  author={Chen, Xi and Bennett, Paul N and Collins-Thompson, Kevyn and Horvitz, Eric},
  booktitle={Proceedings of the sixth ACM international conference on Web search and data mining},
  pages={193--202},
  year={2013}
}

@inproceedings{bunian2021vins,
  title={Vins: Visual search for mobile user interface design},
  author={Bunian, Sara and Li, Kai and Jemmali, Chaima and Harteveld, Casper and Fu, Yun and Seif El-Nasr, Magy Seif},
  booktitle={Proceedings of the 2021 CHI Conference on Human Factors in Computing Systems},
  pages={1--14},
  year={2021}
}

@article{luera2025mllm,
  title={Mllm as a ui judge: Benchmarking multimodal llms for predicting human perception of user interfaces},
  author={Luera, Reuben A and Rossi, Ryan and Dernoncourt, Franck and Basu, Samyadeep and Kim, Sungchul and Mukherjee, Subhojyoti and Mathur, Puneet and Zhang, Ruiyi and Kil, Jihyung and Lipka, Nedim and others},
  journal={arXiv preprint arXiv:2510.08783},
  year={2025}
}

@inproceedings{duan2024generating,
  title={Generating automatic feedback on ui mockups with large language models},
  author={Duan, Peitong and Warner, Jeremy and Li, Yang and Hartmann, Bjoern},
  booktitle={Proceedings of the 2024 CHI Conference on Human Factors in Computing Systems},
  pages={1--20},
  year={2024}
}

@article{duan2024visual,
  title={Visual Prompting with Iterative Refinement for Design Critique Generation},
  author={Duan, Peitong and Cheng, Chin-Yi and Hartmann, Bjoern and Li, Yang},
  journal={arXiv preprint arXiv:2412.16829},
  year={2024}
}

@article{yun2025designlab,
  title={Designlab: Designing slides through iterative detection and correction},
  author={Yun, Jooyeol and Wang, Heng and Shimose, Yotaro and Choo, Jaegul and Takamatsu, Shingo},
  journal={arXiv preprint arXiv:2507.17202},
  year={2025}
}

@inproceedings{lee2019image,
  title={Image aesthetic assessment based on pairwise comparison a unified approach to score regression, binary classification, and personalization},
  author={Lee, Jun-Tae and Kim, Chang-Su},
  booktitle={Proceedings of the IEEE/CVF International Conference on Computer Vision},
  pages={1191--1200},
  year={2019}
}

@inproceedings{goree2023correct,
  title={Correct for whom? subjectivity and the evaluation of personalized image aesthetics assessment models},
  author={Goree, Samuel and Khoo, Weslie and Crandall, David J},
  booktitle={Proceedings of the AAAI Conference on Artificial Intelligence},
  volume={37},
  number={10},
  pages={11818--11827},
  year={2023}
}

@inproceedings{peng2024dreamstruct,
  title={Dreamstruct: Understanding slides and user interfaces via synthetic data generation},
  author={Peng, Yi-Hao and Huq, Faria and Jiang, Yue and Wu, Jason and Li, Xin Yue and Bigham, Jeffrey P and Pavel, Amy},
  booktitle={European Conference on Computer Vision},
  pages={466--485},
  year={2024},
  organization={Springer}
}

@inproceedings{ma2025hpsv3,
  title={Hpsv3: Towards wide-spectrum human preference score},
  author={Ma, Yuhang and Wu, Xiaoshi and Sun, Keqiang and Li, Hongsheng},
  booktitle={Proceedings of the IEEE/CVF International Conference on Computer Vision},
  pages={15086--15095},
  year={2025}
}

@inproceedings{karthik2025scalable,
  title={Scalable ranked preference optimization for text-to-image generation},
  author={Karthik, Shyamgopal and Coskun, Huseyin and Akata, Zeynep and Tulyakov, Sergey and Ren, Jian and Kag, Anil},
  booktitle={Proceedings of the IEEE/CVF International Conference on Computer Vision},
  pages={18399--18410},
  year={2025}
}

@inproceedings{wallace2024diffusion,
  title={Diffusion model alignment using direct preference optimization},
  author={Wallace, Bram and Dang, Meihua and Rafailov, Rafael and Zhou, Linqi and Lou, Aaron and Purushwalkam, Senthil and Ermon, Stefano and Xiong, Caiming and Joty, Shafiq and Naik, Nikhil},
  booktitle={Proceedings of the IEEE/CVF Conference on Computer Vision and Pattern Recognition},
  pages={8228--8238},
  year={2024}
}

@inproceedings{ren2017personalized,
  title     = {Personalized Image Aesthetics},
  author    = {Ren, Jian and Shen, Xiaohui and Lin, Zhe and Mech, Radomir and Foran, David J.},
  booktitle = {Proceedings of the IEEE International Conference on Computer Vision (ICCV)},
  pages     = {638--647},
  year      = {2017},
  doi       = {10.1109/ICCV.2017.76}
}

@article{yang2022para,
  title   = {Personalized Image Aesthetics Assessment with Rich Attributes},
  author  = {Yang, Yuzhe and Xu, Liwu and Li, Leida and Qie, Nan and Li, Yaqian and Zhang, Peng and Guo, Yandong},
  journal = {arXiv preprint arXiv:2203.16754},
  year    = {2022},
  url     = {https://arxiv.org/abs/2203.16754}
}

@article{yun2024taskvector,
  title   = {Scaling Up Personalized Image Aesthetic Assessment via Task Vector Customization},
  author  = {Yun, Jooyeol and Choo, Jaegul},
  journal = {arXiv preprint arXiv:2407.07176},
  year    = {2024},
  url     = {https://arxiv.org/abs/2407.07176}
}

@inproceedings{wang2023agespecificiqa,
  title     = {Age-Specific Perceptual Image Quality Assessment},
  author    = {Wang, Yinan and Chubarau, Andrei and Yoo, Hyunjin and Akhavan, Tara and Clark, James},
  booktitle = {IS{\&}T International Symposium on Electronic Imaging 2023, Image Quality and System Performance XX},
  year      = {2023},
  doi       = {10.2352/EI.2023.35.8.IQSP-302}
}

@article{cherepkova2024contrast,
  title   = {Individual Contrast Preferences in Natural Images},
  author  = {Cherepkova, Olga and Amirshahi, Seyed Ali and Pedersen, Marius},
  journal = {Journal of Imaging},
  volume  = {10},
  number  = {1},
  pages   = {25},
  year    = {2024},
  doi     = {10.3390/jimaging10010025}
}

@article{talebi2018nima,
  title   = {NIMA: Neural Image Assessment},
  author  = {Talebi, Hossein and Milanfar, Peyman},
  journal = {IEEE Transactions on Image Processing},
  volume  = {27},
  number  = {8},
  pages   = {3998--4011},
  year    = {2018},
  doi     = {10.1109/TIP.2018.2831899}
}

@inproceedings{ke2021musiq,
  title     = {MUSIQ: Multi-Scale Image Quality Transformer},
  author    = {Ke, Junjie and Wang, Qifei and Wang, Yilin and Milanfar, Peyman and Yang, Feng},
  booktitle = {Proceedings of the IEEE/CVF International Conference on Computer Vision (ICCV)},
  pages     = {5128--5137},
  year      = {2021}
}

@inproceedings{wang2023clipiqa,
  title     = {Exploring {CLIP} for Assessing the Look and Feel of Images},
  author    = {Wang, Jianyi and Chan, Kelvin C. K. and Loy, Chen Change},
  booktitle = {Proceedings of the AAAI Conference on Artificial Intelligence},
  volume    = {37},
  number    = {2},
  pages     = {2555--2563},
  year      = {2023},
  doi       = {10.1609/aaai.v37i2.25353}
}

@article{caron2012gbt,
  author  = {Caron, Fran{\c{c}}ois and Doucet, Arnaud},
  title   = {Efficient Bayesian Inference for Generalized {Bradley--Terry} Models},
  journal = {Journal of Computational and Graphical Statistics},
  year    = {2012},
  volume  = {21},
  number  = {1},
  pages   = {174--196},
  doi     = {10.1080/10618600.2012.638220}
}

@article{dawid1979ml,
  author  = {Dawid, A. P. and Skene, A. M.},
  title   = {Maximum Likelihood Estimation of Observer Error-Rates Using the EM Algorithm},
  journal = {Journal of the Royal Statistical Society: Series C (Applied Statistics)},
  year    = {1979},
  volume  = {28},
  number  = {1},
  pages   = {20--28}
}

@article{raykar2010crowds,
  author  = {Raykar, Vikas C. and Yu, Shipeng and Zhao, Linda H. and Valadez, Germ{\'a}n H. and Florin, Charles and Bogoni, Luca and Moy, Linda},
  title   = {Learning From Crowds},
  journal = {Journal of Machine Learning Research},
  year    = {2010},
  volume  = {11},
  pages   = {1297--1322}
}

@article{touvron2023llama2,
  title   = {Llama 2: Open Foundation and Fine-Tuned Chat Models},
  author  = {Touvron, Hugo and Martin, Louis and Stone, Kevin and others},
  journal = {arXiv:2307.09288},
  year    = {2023}
}

@inproceedings{song2024veriscore,
  title     = {VERISCORE: Evaluating the Factuality of Verifiable Claims in Long-form Text Generation},
  author    = {Song, Yixiao and Kim, Yekyung and Iyer, Mohit},
  booktitle = {Findings of the Association for Computational Linguistics: EMNLP 2024},
  pages     = {9447--9474},
  year      = {2024}
}

@article{wang2024lrhp,
  title   = {LRHP: Learning Representations for Human Preferences via Preference Pairs},
  author  = {Wang, Chenglong and Gan, Yang and Huo, Yifu and Mu, Yongyu and He, Qiaozhi and Yang, Murun and Xiao, Tong and Zhang, Chunliang and Liu, Tongran and Zhu, Jingbo},
  journal = {arXiv:2410.04503},
  year    = {2024}
}

@inproceedings{peng2025morae,
  title={Morae: Proactively Pausing UI Agents for User Choices},
  author={Peng, Yi-Hao and Li, Dingzeyu and Bigham, Jeffrey P and Pavel, Amy},
  booktitle={Proceedings of the 38th Annual ACM Symposium on User Interface Software and Technology},
  pages={1--14},
  year={2025}
}

@inproceedings{peng2021slidecho,
  title={Slidecho: Flexible non-visual exploration of presentation videos},
  author={Peng, Yi-Hao and Bigham, Jeffrey P and Pavel, Amy},
  booktitle={Proceedings of the 23rd International ACM SIGACCESS Conference on Computers and Accessibility},
  pages={1--12},
  year={2021}
}

@inproceedings{qiao2025thematic,
  title={Thematic-LM: a LLM-based multi-agent system for large-scale thematic analysis},
  author={Qiao, Tingrui and Walker, Caroline and Cunningham, Chris and Koh, Yun Sing},
  booktitle={Proceedings of the ACM on Web Conference 2025},
  pages={649--658},
  year={2025}
}

@article{wang2025lata,
  title={LATA: A Pilot Study on LLM-Assisted Thematic Analysis of Online Social Network Data Generation Experiences},
  author={Wang, Qile and Erqsous, Moath and Barner, Kenneth E and Mauriello, Matthew Louis},
  journal={Proceedings of the ACM on Human-Computer Interaction},
  volume={9},
  number={2},
  pages={1--28},
  year={2025},
  publisher={ACM New York, NY, USA}
}

@inproceedings{peng2021say,
  title={Say it all: Feedback for improving non-visual presentation accessibility},
  author={Peng, Yi-Hao and Jang, JiWoong and Bigham, Jeffrey P and Pavel, Amy},
  booktitle={Proceedings of the 2021 CHI Conference on Human Factors in Computing Systems},
  pages={1--12},
  year={2021}
}

@inproceedings{beeferman2023feedbackmap,
  title={FeedbackMap: A tool for making sense of open-ended survey responses},
  author={Beeferman, Doug and Gillani, Nabeel},
  booktitle={Companion Publication of the 2023 Conference on Computer Supported Cooperative Work and Social Computing},
  pages={395--397},
  year={2023}
}

@inproceedings{lambert2025rewardbench,
  title={Rewardbench: Evaluating reward models for language modeling},
  author={Lambert, Nathan and Pyatkin, Valentina and Morrison, Jacob and Miranda, Lester James Validad and Lin, Bill Yuchen and Chandu, Khyathi and Dziri, Nouha and Kumar, Sachin and Zick, Tom and Choi, Yejin and others},
  booktitle={Findings of the Association for Computational Linguistics: NAACL 2025},
  pages={1755--1797},
  year={2025}
}

@article{malik2025rewardbench2,
  title={RewardBench 2: Advancing Reward Model Evaluation},
  author={Malik, Saumya and Pyatkin, Valentina and Land, Sander and Morrison, Jacob and Smith, Noah A and Hajishirzi, Hannaneh and Lambert, Nathan},
  journal={arXiv preprint arXiv:2506.01937},
  year={2025}
}

@inproceedings{hsu2025postero,
  title={PosterO: Structuring Layout Trees to Enable Language Models in Generalized Content-Aware Layout Generation},
  author={Hsu, HsiaoYuan and Peng, Yuxin},
  booktitle={Proceedings of the Computer Vision and Pattern Recognition Conference},
  pages={8117--8127},
  year={2025}
}

@inproceedings{peng2023slide,
  title={Slide gestalt: Automatic structure extraction in slide decks for non-visual access},
  author={Peng, Yi-Hao and Chi, Peggy and Kannan, Anjuli and Morris, Meredith Ringel and Essa, Irfan},
  booktitle={Proceedings of the 2023 CHI Conference on Human Factors in Computing Systems},
  pages={1--14},
  year={2023}
}

@inproceedings{hsu2023posterlayout,
  title={Posterlayout: A new benchmark and approach for content-aware visual-textual presentation layout},
  author={Hsu, Hsiao Yuan and He, Xiangteng and Peng, Yuxin and Kong, Hao and Zhang, Qing},
  booktitle={Proceedings of the IEEE/CVF Conference on Computer Vision and Pattern Recognition},
  pages={6018--6026},
  year={2023}
}

@inproceedings{ge2025autopresent,
  title={Autopresent: Designing structured visuals from scratch},
  author={Ge, Jiaxin and Wang, Zora Zhiruo and Zhou, Xuhui and Peng, Yi-Hao and Subramanian, Sanjay and Tan, Qinyue and Sap, Maarten and Suhr, Alane and Fried, Daniel and Neubig, Graham and others},
  booktitle={Proceedings of the Computer Vision and Pattern Recognition Conference},
  pages={2902--2911},
  year={2025}
}

@inproceedings{peng2022diffscriber,
  title={Diffscriber: Describing visual design changes to support mixed-ability collaborative presentation authoring},
  author={Peng, Yi-Hao and Wu, Jason and Bigham, Jeffrey and Pavel, Amy},
  booktitle={Proceedings of the 35th Annual ACM Symposium on User Interface Software and Technology},
  pages={1--13},
  year={2022}
}

@misc{li2025waybackui,
  title  = {WaybackUI: A Dataset to Support the Longitudinal Analysis of Web User Interfaces},
  author = {Li, Amanda and Peng, Yi-Hao and Nichols, Jeff and Bigham, Jeff and Wu, Jason},
  year   = {2025},
  note   = {arXiv, 2025.12}
}

@inproceedings{wu2023webui,
  title={Webui: A dataset for enhancing visual ui understanding with web semantics},
  author={Wu, Jason and Wang, Siyan and Shen, Siman and Peng, Yi-Hao and Nichols, Jeffrey and Bigham, Jeffrey P},
  booktitle={Proceedings of the 2023 CHI Conference on Human Factors in Computing Systems},
  pages={1--14},
  year={2023}
}

@String(ICCV= {Int. Conf. Comput. Vis.})

@String(AAAI = {AAAI})

@String(ICCV  = {ICCV})
}





\clearpage
\onecolumn                
\setcounter{page}{1}

\begin{center}
  {\Large \textbf{\thetitle}}\\[4pt]
  {\large Supplementary Material}
\end{center}

\setcounter{section}{0}
\renewcommand{\thesection}{S\arabic{section}}

\section{Model Hyperparameters}
\label{sec:parameters}

Here are the hyperparameteres we used for training and prompting VLMs:

\begin{table}[htbp]
\centering
\caption{Training hyperparameters for \textsc{UIClip}.}
\label{tab:hyperparams-uiclip}
\begin{tabular}{ll}
\toprule
\textbf{Model} & \textsc{UIClip} \\
\midrule
\textbf{Learning rate} & $5 \times 10^{-4}$ \\
\textbf{Schedule} & Cosine ($T_{\max}=24$, $\eta_{\min}=10^{-4}$) \\
\textbf{Weight decay} & $10^{-2}$ \\
\textbf{Batch size} & 64 \\
\textbf{Gradient clipping} & 1.0 \\
\textbf{Patience} & 5 \\
\textbf{Margin multiplier} & 1.1 for ``much better'' labels \\
\bottomrule
\end{tabular}
\end{table}

\noindent All the LMM's parameters for randomness or reasoning are set to default.

\section{Prompt Sets}
\label{sec:prompt}

\begin{tcolorbox}[
  colback=orange!5,
  colframe=orange!80!black,
  title={\textbf{Visual UI Design Generation}},
  fonttitle=\bfseries,
  coltitle=white,
  boxrule=0.6pt,
  arc=2mm,
  left=5pt,
  right=5pt,
  top=4pt,
  bottom=4pt,
  before upper={\parindent0pt\raggedright}
]
\small
\textbf{You are an expert mobile UI designer and developer.}

\medskip
\textbf{Goal}
\begin{itemize}
  \item Output ONE standalone HTML file that renders ONE static mobile screen.
\end{itemize}

\medskip
\textbf{Canvas}
\begin{itemize}
  \item Wrap all content in\\
        \texttt{<div class="ui-canvas">…</div>}.
  \item Size: 628x1118 px. Center the canvas on the page.
  \item No horizontal or vertical scroll. No overlap or cropping. Keep all content inside \texttt{.ui-canvas}.
  \item Include a small guard style to center the canvas and hide page overflow.
\end{itemize}

\medskip
\textbf{Tech}
\begin{itemize}
  \item Use only HTML, CSS, and JavaScript. No build tools.
  \item Use Tailwind via CDN in \texttt{<head>}:
        $<$script src="https://cdn.tailwindcss.com"$>$$<$/script$>$.
  \item If you use icons, import the icon library via CDN before use (e.g., Heroicons) or inline SVG.
  \item If you add charts or 3D, prefer D3/Recharts/Three.js via CDN U.
\end{itemize}

\medskip
\begin{itemize}
  \item Use semantic tags. Bind labels to controls with \texttt{for}/\texttt{id}.
  \item Provide meaningful alt text for images.
  \item Meet WCAG criteria for color contrast and content style. Show visible \texttt{:focus-visible} outlines.
  \item Group radios by \texttt{name}. Switches use native checkbox under the hood; reflect checked/disabled; label is clickable.
  \item Sliders use \texttt{<input type="range">} with a label and a live value.
  \item Tabs or segmented controls show selected state and support keyboard focus.
  \item If a modal exists, trap focus, provide ESC close, and a close button.
\end{itemize}

\medskip
\textbf{Images (Assets)}
\begin{itemize}
  \item If the user provides asset URLs, use ONLY those URLs in \texttt{<img src="...">}. Provide concise, specific alt text.
  \item Do not reference Unsplash/Picsum/placeholder services. Do not leave empty \texttt{src} attributes.
\end{itemize}

\medskip
\textbf{Return}
\begin{itemize}
  \item Return ONLY valid HTML from \texttt{<!DOCTYPE html>} to \texttt{</html>}. No commentary or Markdown.
\end{itemize}
\end{tcolorbox}

\begin{tcolorbox}[
  colback=blue!5,
  colframe=blue!80!black,
  title={\textbf{UI Preference Judging Prompts}},
  fonttitle=\bfseries,
  coltitle=white,
  boxrule=0.6pt,
  arc=2mm,
  left=5pt,
  right=5pt,
  top=4pt,
  bottom=4pt,
  before upper={\parindent0pt\raggedright}
]
\small
\textbf{(A) Zero-shot UI Judge}

\textit{Developer instructions.}
The model rates UIs for a generic population:
\begin{itemize}
  \item Choose the screen most people would prefer.
  \item Use a rubric over clarity, readability, spacing and alignment, emphasis of primary actions,
        aesthetic balance, and fit to the screen prompt.
  \item Evaluate only visible features, avoid position bias, do not allow ties,
        and keep reasons short and concrete.
  \item Print ONLY the fixed output template.
\end{itemize}

\textit{User message.}
The user provides one screen prompt (optional) and two images (Image A and Image B), then the output template:
\begin{itemize}
  \item \texttt{CHOICE\_4WAY: <A >> B | A > B | B > A | B >> A>}
  \item \texttt{BINARY\_PREFERENCE: <A | B>}
  \item \texttt{CONFIDENCE: <0.00-1.00>}
  \item \texttt{REASONS:} with 2--3 short bullet points.
\end{itemize}

\medskip
\textbf{(B) Few-shot UI Judge (Personalized or Pooled)}

\textit{Developer instructions.}
The model adapts to one user's taste:
\begin{itemize}
  \item Infer this user's preference pattern from labeled examples and apply it to new pairs.
  \item When the general rubric disagrees with the user's past choices, follow the user's taste.
  \item Use the same rubric only as a tie breaker, and obey the same rules on visibility, bias, ties,
        and concise reasons as above.
\end{itemize}

\textit{User message with examples.}
The prompt first lists several labeled examples for that user:
\begin{itemize}
  \item For each example: \texttt{Example i (user-labeled)}, optional \texttt{Screen prompt},
        images A and B, and \texttt{User's label: <A >> B | A > B | B > A | B >> A>} (score in $\{-2,-1,1,2\}$).
\end{itemize}

Then the prompt introduces the target pair:
\begin{itemize}
  \item \texttt{Now predict this user's preference for the TARGET pair.}
  \item \texttt{Screen prompt}, images A and B, and the same fixed output template:
        \texttt{CHOICE\_4WAY}, \texttt{BINARY\_PREFERENCE}, \texttt{CONFIDENCE}, and \texttt{REASONS}.
\end{itemize}

\end{tcolorbox}

\section{Guidelines for Preference Annotations}
\label{sec:guidelines}

\begin{tcolorbox}[
  colback=cyan!5,
  colframe=teal!80!black,
  title=\textbf{Preference Ratings Labeling (1 out of 4 options)},
  fonttitle=\bfseries,
  coltitle=white,
  boxrule=0.6pt,
  arc=2mm,
  left=5pt,
  right=5pt,
  top=4pt,
  bottom=4pt
]
\begin{enumerate}
\item Based on your personal preferences, click the option or use 1,2,3,4 to select the preferences
1: $A >> B, 2: A > B, 3: A < B, 4: A << B$

\item Use left and right to go forward and backward

\item Focus less on the quality of the placeholder content (e.g., generated text) and more on the overall design (layout, color choices, style ..etc); Sometimes some images may not even be loaded properly. Instead of focusing on the unloaded assets, focus on the overall design

\item If you feel both of them are of equal quality because they both contain flaws, think about 
which one you would rather have as a starting point if it was your job to fix it.
\end{enumerate}
\end{tcolorbox}

\begin{tcolorbox}[
  colback=cyan!5,
  colframe=teal!80!black,
  title=\textbf{Rationale Specification for Preference Choices},
  fonttitle=\bfseries,
  coltitle=white,
  boxrule=0.6pt,
  arc=2mm,
  left=5pt,
  right=5pt,
  top=4pt,
  bottom=4pt
]
\begin{enumerate}
\item For each pair of UIs, you see your previous preference from Study~1. You may change the choice by clicking a button or pressing keys 1--4.

\item Refer to the designs as ``screen~A'' and ``screen~B'' (or ``image~A'' and ``image~B''). You may also write simply ``A'' and ``B'' in your explanation.

\item Write a rationale of at least 2--3 sentences for every choice. Explain why you prefer one screen and why you chose that strength of preference.

\item Give concrete visual and UI reasons, such as layout, color, hierarchy, typography, spacing, alignment, or missing components from the prompt. Avoid short generic statements like ``A looks better than B'' without details.

\item If both screens have flaws, choose the one you would prefer as a starting point to fix and explain why. Very short or vague rationales may be rejected and may not receive compensation.
\end{enumerate}
\end{tcolorbox}

\clearpage

\end{document}